\documentclass[10pt,twocolumn,letterpaper]{article}

\usepackage{iccv}
\usepackage{times}
\usepackage{epsfig}
\usepackage{graphicx}
\usepackage{amsmath}
\usepackage{amssymb}
\usepackage{multirow}
\usepackage{adjustbox}
\usepackage{xcolor}
\usepackage{booktabs}
\usepackage{tikz}
\usepackage{pgfplots}
\usepackage{pgfplotstable}
\usepackage{caption}
\usepackage{subcaption}
\usepackage{amssymb}
\usepackage{pifont}

\definecolor{red1}{HTML}{EA484B}
\definecolor{red-light}{HTML}{f4a4a5}
\definecolor{yellow1}{HTML}{FDAE61}
\definecolor{yellow-light}{HTML}{fdcc9b}
\definecolor{green1}{HTML}{19c20a}
\definecolor{green-light}{HTML}{a6fa9e}
\definecolor{blue1}{HTML}{2B83BA}
\definecolor{blue-light}{HTML}{acd3ec}
\definecolor{orange1}{HTML}{ff9900}

\usepackage[breaklinks=true,bookmarks=false]{hyperref}

\iccvfinalcopy 

\def\iccvPaperID{9944} 

\ificcvfinal\fi

\begin{document}

\title{SATA: Source Anchoring and Target Alignment Network for Continual Test Time Adaptation}

\author{Goirik Chakrabarty\\
IISER, Pune\\
{\tt\small goirik.chakrabarty@students.iiserpune.ac.in}
\and
Manogna Sreenivas \quad Soma Biswas\\
IISc, Bangalore\\
{\tt\small {\{manognas, somabiswas\}}@iisc.ac.in}
}

\maketitle
\ificcvfinal\thispagestyle{empty}\fi

\begin{abstract}
Adapting a trained model to perform satisfactorily on continually changing testing domains/environments is an important and challenging task. 
In this work, we propose a novel framework, SATA, which aims to satisfy the following characteristics required for online adaptation: 
1) can work seamlessly with different (preferably small) batch sizes to reduce latency; 
2) should continue to work well for the source domain;
3) should have minimal tunable hyper-parameters and storage requirements.
Given a pre-trained network trained on source domain data, the proposed SATA framework modifies the batch-norm affine parameters using source anchoring based self-distillation.
This ensures that the model incorporates the knowledge of the newly encountered domains, without catastrophically forgetting about the previously seen ones. 
We also propose a source-prototype driven contrastive alignment to ensure natural grouping of the target samples, while maintaining the already learnt semantic information. 
Extensive evaluation on three benchmark datasets under challenging settings justify the effectiveness of SATA for real-world applications.
\end{abstract}

\section{Introduction} 

Deep learning has achieved phenomenal success in several computer vision tasks like classification, object detection, segmentation, etc \cite{imagenet, faster_rcnn, mask_rcnn, deeplab, pascal-voc}.
But it is well known that these models tend to perform poorly when the test data comes from a different distribution compared to the training data~\cite{ovadia2019can}.
Unsupervised Domain Adaptation (UDA) techniques~\cite{dan, mcd} have shown promising results in this scenario, but they require access to unlabaled target data along with the labelled source data, which may be difficult due to privacy concerns, storage constraints, etc.
In addition to having a different distribution compared to the training data, the test data distribution may dynamically vary with time. 
For example, in autonomous driving, the model trained using data taken in clear weather, can encounter cloudy weather followed by heavy rain, etc. during deployment. 
Thus, continually adapting the model during test-time is critical for the model to perform well in changing scenarios.

Test-time adaptation of trained models has thus emerged as an important research area, where an off-the-shelf trained model is adapted to the testing data as and when they are encountered. 
The majority of successful frameworks for this task, \cite{tent, lame}, assume that the test data belongs to a single domain, which is a restrictive assumption for practical applications. 
Very recently, researchers have started to look at the continual test time adaptation setting~\cite{cotta}, where the target distribution can change over time. 

In this work, we propose a novel framework, termed {\bf S}ource {\bf A}nchored and {\bf T}arget {\bf A}lignment (SATA) Network, for the task of continual test-time domain adaptation. 
We feel that for the model to be practically useful in an online setting, it should satisfy the following requirements:
{\em
1) For online adaptation, the models should work seamlessly with different (preferably small) batch sizes which reduces the inference time and latency;
2) The updated model should continue to work well on the source domain;
3) The framework should require less storage and minimal tunable hyper-parameters, since validation sets are usually not available during test-time.}
With this motivation, we propose to use source anchoring based self-distillation, 
which ensures that the model robustly adapts to the incoming data, while not forgetting the source domain information.
The proposed SATA also utilizes contrastive learning to ensure better model generalizability to unseen domains. 
Here, we also utilize the source prototypes for alignment of the target features to the corresponding source data, which help to conserve the semantic information learnt using the source.
We propose to only update the BN affine parameters like TENT~\cite{tent}, which helps to avoid overfitting on the small amount of target data, in addition to reducing the storage requirements. 
This simple, yet effective framework helps us take a step forward in achieving all the objectives mentioned earlier. 
Extensive experiments on three large-scale benchmark datasets, namely CIFAR-10C, CIFAR-100C and ImageNet-C \cite{imagenetc} for different challenging and realistic scenarios justify the effectiveness of the proposed SATA framework. 
To summarize, the contributions of this work are as follows:
\begin{itemize}
    \item We propose a novel SATA framework for the task of continual test-time domain adaptation.
    \item The proposed framework takes a step forward in overcoming some of the important challenges in a practical test-time adaptation setting. 
    \item We show that the proposed source-based anchoring along with the source-guided contrastive alignment can be successfully utilized for robustly updating the model under dynamically changing test conditions.
    \item Extensive evaluation on challenging settings justifies its effectiveness for different scenarios.
\end{itemize}
We now discuss the related work, followed by the proposed method and evaluation.


\section{Related works}
The proposed approach is inspired by several seminal works in different areas.
Here, we provide pointers to some of the related work in literature. \\ \\
{\bf Source Free Domain Adaptation:} Source-free domain adaptation methods aim to adapt a model trained using source data to a new domain using only the unlabeled target data~\cite{sfda-a2net-xia2021adaptive, sfda-hcl-huang2021model, sfda-nrc-yang2021exploiting, sfda-shot-liang2020we, sfda1-kundu2020universal, sfda2-li2020model, sfda3-tian2021vdm, gsfda-yang2021generalized}, 
Some of these methods, such as SHOT~\cite{sfda-shot-liang2020we}, use a pseudo-labelling strategy to maximize information and minimize entropy, while others, such as~\cite{sfda2-li2020model}, use generative models to enhance model performance on the target domain by generating target-style images. 
GSFDA~\cite{gsfda-yang2021generalized} aims to activate different channels within a network for different domains while also taking into account the local data structure. 
A$^2$-Net~\cite{sfda-a2net-xia2021adaptive} is another method that utilizes different classifiers to align the two domains using adversarial training. \\ \\
{\bf (Continual) Test time adaptation: }
Test-time Adaptation (TTA) is an online variant of SFDA that adapts the model at test time using small batches of test data, as and when they become available.
Here, the model's parameters or architecture are usually adjusted to handle the differences between the two domains better, thereby improving its performance on the target domain \cite{tent, pseudo, lame, tta-autoencoder, bn_neurips}. 
Some of these methods focus on modifying the original architecture during the source training like TTT~\cite{ttt}, which trains the model on supervised and self-supervised tasks using source data. 
During testing, the self-supervised module is fine-tuned on the target data to improve performance. 
Recently, several researchers are focusing on the fully test time adaptation setting~\cite{tent}, \cite{bn_neurips}, which does not assume any access to source data or the source training process making it more practical.
TENT~\cite{tent} adopts entropy minimization objective for training the BN layers, while BNStatsAdapt~\cite{bn_neurips} adjusts the BN statistics during test time to align the target with the source domain.

A more realistic scenario is handled by the recently proposed continual test-time adaptation protocol~\cite{cotta}, where the trained model should continually adapt to a dynamic environment, where the test domain can change over time.
CoTTA~\cite{cotta} utilizes weight-averaged and augmentation-averaged predictions to reduce error accumulation and also stochastically restores a small part of the neurons to the source pre-trained weights during each iteration to avoid catastrophic forgetting. 
This allows for long-term adaptation of all parameters in the network while preserving source knowledge.
Recently, several modules have been developed which can aid the dynamic adaptation to test-data. 
However, these modules need to be optimised along with the model during training with source~\cite{newcomers, note}. \\ \\
{\bf Knowledge and Self-distillation:} Knowledge distillation is a technique used to transfer knowledge from a large, complex model ("teacher" model) to a smaller, simpler model ("student" model) \cite{hinton2015distilling, kd-survey}. 
This is done by training the student model to mimic the predictions of the teacher model, which has already learned useful representations, rather than training the student model on the original labeled data.
CoTTA also utilizes distillation method to enhance the adaptation to new domains, which involves the implementation of a teacher model to make accurate predictions based on the student model.


\begin{figure*}[t]
    \centering
    \includegraphics[width=\textwidth]{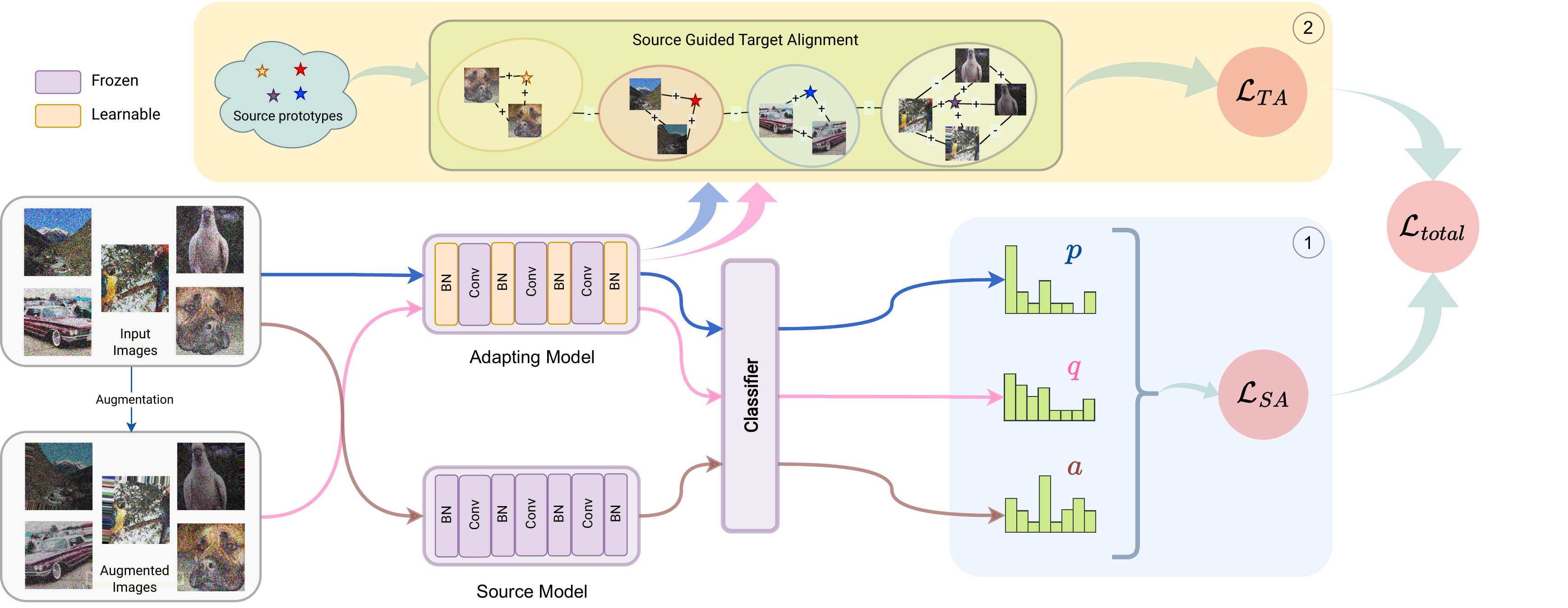}
    \caption{
Illustration of the proposed SATA framework. The original image and its augmentation are passed through the adapting model (only BN affine parameters are updated) and the source model. \ding{192} The prediction of the input given by the source model is used as an anchor ($\mathbf{a}$) for the prediction given by the adapting model ($\mathbf{p}$ for input image and {$\mathbf{q}$} for augmented image) for computing the source-anchoring loss $\mathcal{L}_{SA}$. \ding{193} The source guided target alignment $\mathcal{L}_{TA}$ uses the source prototypes to help maintain the semantic information learnt using the source domain and enforces clustering that is meaningful in the feature space. The augmented input is also used for both the losses. }
    \label{fig:my_label}
\end{figure*}

A variant termed as self-distillation or self-knowledge distillation \cite{kd-self}, involves training a model to mimic its own predictions. 
The framework in~\cite{yang2019snapshot} shows that using a model from a previous epoch to train the same model in future epochs can increase the training efficiency and accuracy of the model. \\ \\
{\bf Contrastive learning:} 
Contrastive learning \cite{simclr, supcon, swav} has shown tremendous improvement in learning visual features for various downstream tasks. Because of its robustness researchers have used it in SFDA~\cite{sfda-hcl-huang2021model, sfda-a2net-xia2021adaptive, sfda-con-cross, yang2022attracting, sfda-con-zhao2022adaptive}, TTA~\cite{adacontrast}, Domain Generalisation (DG) \cite{dom-gen-con1, dom-gen-con-pcl} and other studies \cite{vaze2022gcd, fei2022xcon} to train/adapt a pre-trained model to a target domain using only unlabeled data. 
In previous works, contrastive learning has been used to get better feature representation. However, in this work, we align target features guided by source prototypes to get meaningful target features with respect to the source feature space.
This results in meaningful clustering and good separation of classes in unseen domains. 

\section{Problem Setting and Motivation}
In continual test-time domain adaptation, we are given a model trained using source domain data $\mathcal{D}_s = \{(x_i, y_i)\}_{i=1}^{n_s} \sim P_s$. Here, $(x_i, y_i)$ is source data and label which is drawn from the source distribution $P_s$. During testing, the source data is usually not available due to privacy concerns or storage constraints. At this stage, the model encounters test data $\mathcal{D}_t = \{x_j\}_{j=1}^{n_t} \sim P_t$. 
In practice, the test data can belong to a different domain compared to the source, i.e. $P_t \neq P_s$. Further, $P_t$ can change over time such that $P_t^{(1)} \neq P_t^{(2)} \neq \hdots \neq P_s$ leading to our continual test time adaptation scenario.


In this work, the goal is to develop a practically useful framework for the above continual fully test time scenarion, which should have the following characteristics:
1) For online adaptation, to reduce the latency, the model should be able to work seamlessly for different (preferably small) batch sizes, like 25 or 10 as used in \cite{small-batch-delta, ttn}. 
But the current approaches~\cite{cotta} have been tested only for much larger batch sizes (like 200);
2) The updated model should continue to work well on the source domain; 
For example, while adapting to cloudy or rainy weather, the framework deployed for autonomous driving should not fail for images taken in clear weather (source domain); 
3) In test-time adaptation scenario, since validation sets are usually not available for tuning hyper-parameters, the approach should require minimal hyper-parameters; and as a practical bonus, storage should be minimal and the framework should have low inference time for ease of deployment.
Now, we describe the proposed SATA framework (Figure~\ref{fig:my_label}), and elaborate on how some of these challenges are addressed.
\section{Proposed SATA Framework}
Given the model trained using the source data, the goal is to adapt it using the limited amount of test-data encountered in each batch in a dynamic environment, while satisfying the desirable criteria mentioned above.
Towards this goal, we propose using (i) Source Anchorization and (ii) Source-Guided Contrastive Alignment, which we now describe.




\subsection{Source Anchorization of Model}

During test-time, the model has access to few test samples in a batch, which may not be representative of the corresponding target distribution.
Thus, modifying the model parameters completely on the basis of the available target data may result in simultaneously overfitting on the few target samples and also catastrophic forgetting of the source information.
CoTTA~\cite{cotta} addressed this challenge by (i) learning a teacher model by combining the source and a continuously adapting student model, wherein the teacher changes gradually for robust prediction and also using (ii) stochastic restoration to reset some of the model parameters to the source model after every batch.
Though this gives impressive performance, it has two limitations, namely 
(i) the complete teacher, student and source model needs to be stored and 
(ii) the hyperparameters required for computing the teacher model and also for stochastic restoration need to be determined, which can have different optimal values for different datasets. 
To overcome these challenges, in this work, we use self-distillation using the source model as the anchor~\cite{hinton2015distilling}\cite{yang2019snapshot}.

We denote the adapting model as $f_{\theta}$, as this is the model which is constantly updated and is used to predict the target data. 
The weights $\theta$ of the adapting model is initialised to the weights $\theta_s$ of the off-the-shelf source model given by $f_{\theta_s}$. 
Now, consider time instant $k$, when the model encounters a new batch denoted by $B_k$.
Since adaptating the BN statistics has proven to be effective in capturing the data distribution characteristics \cite{bn_neurips}\cite{cotta}, the BN statistics of the source model ($f_{\theta_s}$) and the adapting model ($f_{\theta}$) are changed to the BN statistics of the target batch at each step. 
Let these models be denoted as $f^k_{\theta_s}$ and $f^k_{\theta}$  respectively. 
$f^k_{\theta_s}$ (referred to as source from now) can be thought of as a specialized model that accounts for the domain difference between the source and the specific target batch, 
On the other hand, the adapting model's weights are optimized after every batch using the loss function that will be described later. 
It should also be noted that during optimization, only the BN parameters are updated for the adapting model~\cite{tent}. 

The proposed self-distillation loss is inspired from the knowledge distillation loss formulation used in incremental learning \cite{few-know-distl, cil-know-distl} to prevent catastrophic forgetting.
In this work, self-distillation between the adapting model and the source model acts as a regularizer~\cite{self-know-distl}, that encourages the adapting model to mimic the source model, which is a specialized model whose response corresponds to domain invariant features. Thus, our adapting model is reinforced to learn domain invariant features~\cite{self-know-distl}, leading to better generalization, which we empirically observe in Table \ref{table:generalization}.

The loss function is based on the prediction scores of the adapting model and source model for a given batch of test images, $B_k:= \{x_1, x_2, \hdots, x_{N_k}\}$. 
Let $p_{ij}$ and $a_{ij}$ denote the $j^{th}$ element of $f^k_{\theta}(x_i)$ (adapting model) and $f^k_{\theta_s}(x_i)$ (source model) respectively, which gives the prediction score of the $j^{th}$ class for the $i^{th}$ test image.
The source-anchoring loss for a given batch is calculated as follows:
\begin{equation}
\label{eq:1}
    \mathcal{L}'_{\textrm{SA}} (B_k) = -\frac{1}{N_k}\sum_{i=1}^{N_k}\sum_{j=1}^{C} p_{ij} \log(a_{ij}) 
\end{equation}
Here, $C$ is the number of classes, and $N_k$ is the number of samples in the $k^{th}$ batch.
Empirically, we observe that using augmentations of the test images make the model more robust.
Let $q_{ij}$ denote the prediction score of the $j^{th}$ class for the $i^{th}$ augmented test image, given by the adapting model. The complete source-anchoring loss is given by 
\begin{equation}
    \mathcal{L}_{\textrm{SA}} (B_k) = -\frac{1}{N_k}\sum_{i=1}^{N_k}\sum_{j=1}^{C} (p_{ij} \log(a_{ij}) + q_{ij} \log(a_{ij}))
\end{equation}
This simple, yet effective self-distillation offers multiple advantages as follows:
i) Since the modified source is used for anchoring, there is no hyper-parameter involved (like weights for combining student with source to form the teacher model);
ii) The adapting model can be directly used for prediction continuously, without requiring any restoration to the source model;
iii) Since only the BN parameters are updated, just these parameters of the source need to be stored, resulting in much lesser storage requirements compared to storing two different models. 

\subsection{Source-Guided Target Alignment}

The goal is to learn a generalized model as it encounters data from different domains, thereby making the features gradually domain invariant. 
Here, we additionally use self-supervision (in the form of contrastive learning) for improved generalization as used in \cite{sfda-con-zhao2022adaptive, sfda-con-cross}.
Formally, the adapting model $f^k_{\theta}$ can be decomposed into a feature extractor, $g^k_\phi$ and the fixed classifier $h$, i.e.
\begin{equation}
    f^k_\theta = h \circ g^k_\phi
\end{equation}
Suppose the augmented samples for the test batch data $\{x_i, y_i\}_{i=1}^{N_k}$ be denoted as $\{x_i, y_i\}_{i=N_k+1}^{2N_k}$ where $y_i = y_{N_k+i}$. 
We use the same augmentations for our experiments as in CoTTA~\cite{cotta}.
These features are then passed to a projection head $p_\psi$ so that the features are mapped to a $d$-dimensional hyper-sphere \cite{simclr, supcon}.
\begin{equation}
    z_i = p_\psi \circ g^k_\phi (x_i)
\end{equation}
Now, the parameters $\psi$ and $\phi$ are optimised using the contrastive loss given below:
\begin{equation}
    \mathcal{L}_{con} = \sum_{i=1}^{2N_k} \frac{-1}{|S_{-i}|} \sum_{j\in S_{-i}} \log \left(\frac{\exp(z_i . z_j/\tau)}{\sum_{k\neq i}\exp(z_i .z_k/\tau)}\right)
\end{equation}
here $\tau > 0$  is the temperature hyperparameter and 
\begin{equation}
S_{-i} = \{j \,|\, j \neq i, y_i = y_j \forall j \in \{1, \hdots, 2N_k\}\}
\end{equation}
In this work, as in~\cite{tent}, only the BN layers of the feature extractor are modified to account for the changing distribution, but the classifier layer remains unchanged (to avoid overfitting on the few target samples in each batch).
This ensures that the semantic information in the feature embedding space is not disturbed during the adaptation process.
Thus, for correct classification, the target clusters should also align with the original source representations, which is achieved using the source-guided alignment loss.
To this end, we include a third view which assigns the nearest source prototype features as an augmented view for the test time features. 
Specifically, let the source prototypes for the $C$ classes be denoted as $\{\pi_i\}_{i=1}^C$.
The source prototype views given by $\{x_i, y_i\}_{i=2N_k+1}^{3N_k}$, such that $y_i = y_{2N_k+i}$ are calculated as follows:

\begin{equation}
    g^k_\phi(x_{2N_k+i}) = \{\pi_j | \arg\max_j(CosineSim(\pi_j, g^k_\phi(x_i)))\} 
\end{equation}

\begin{figure}[]
    \centering
    \begin{subfigure}[b]{0.23\textwidth}
        \centering
        \includegraphics[width=\textwidth]{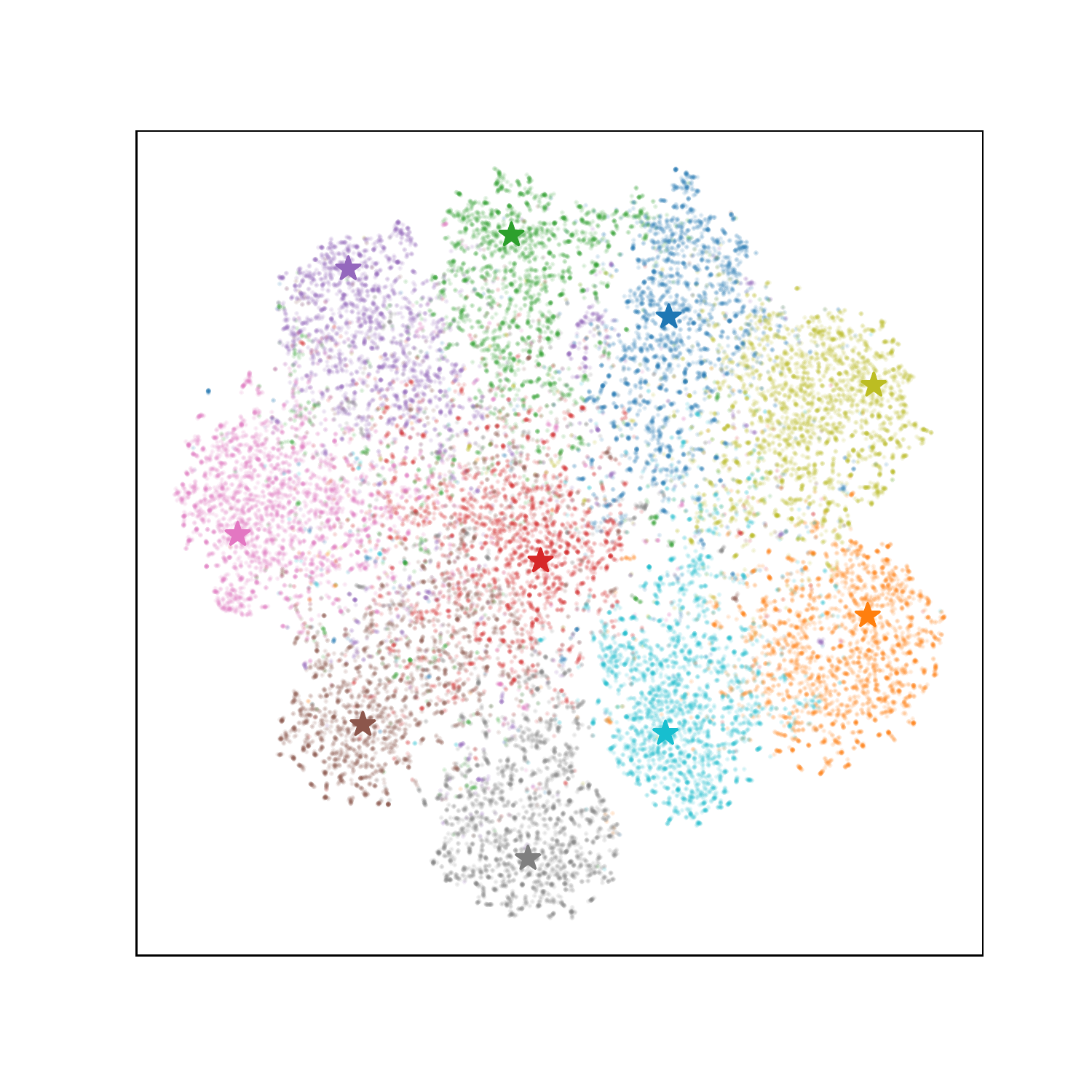}
        \caption{}
    \end{subfigure}
     \hfill
    \begin{subfigure}[b]{0.23\textwidth}
        \centering
        \includegraphics[width=\textwidth]{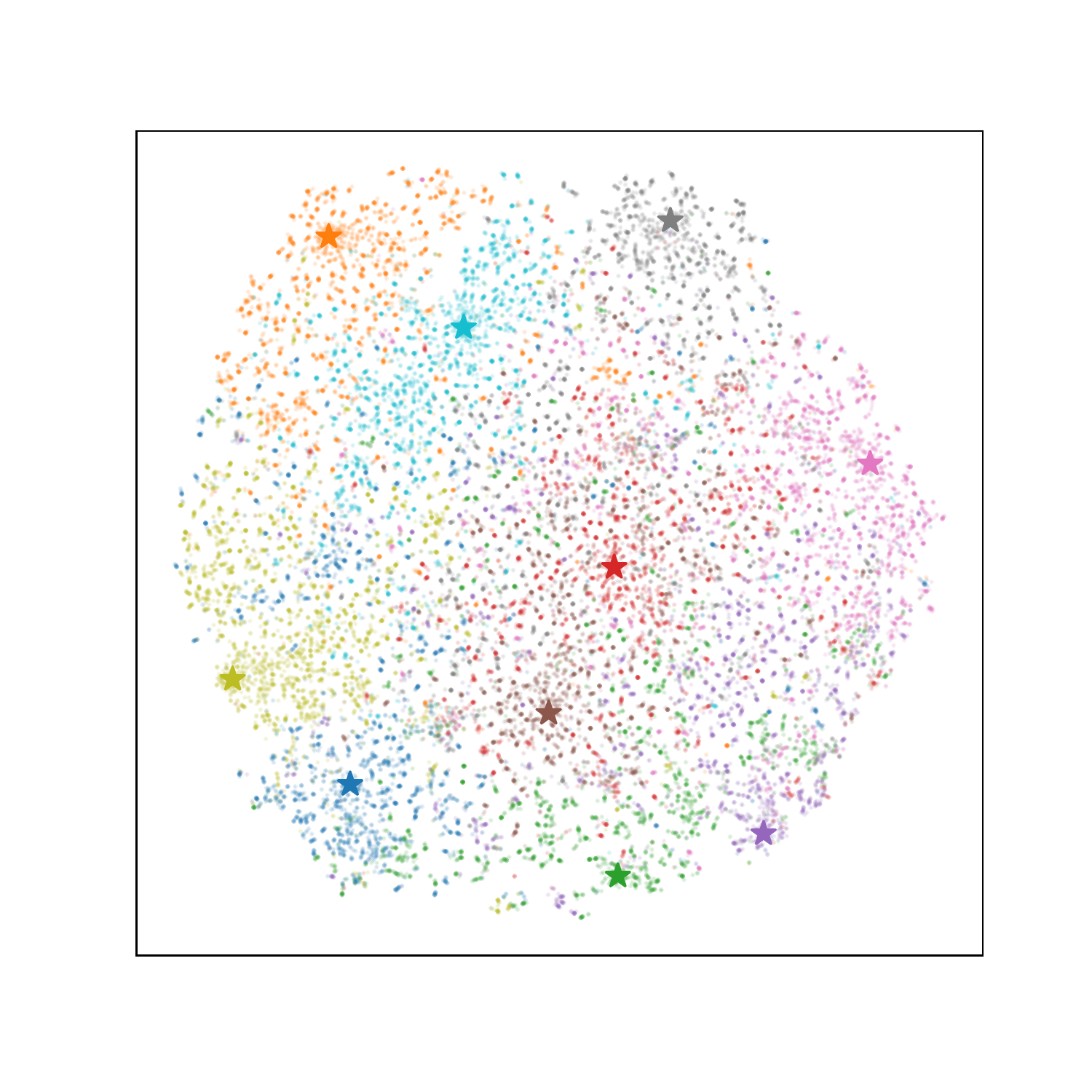}
        \caption{}
    \end{subfigure}
     \hfill
   \caption{t-SNE plot of the feature space using only $\mathcal{L}_{\textrm{TA}}$ for adaptation.
   We observe that the alignment and clustering is better with source prototypes (a) as compared to not using them (b). The source prototypes are represented using stars.}
   \label{fig:TA}
\end{figure}

\noindent The effect of using prototypes as a view can be seen in Figure \ref{fig:TA}. We see that using this view (Eq 7 directly gives features) implicitly passes class information to the contrastive learning algorithm resulting in improved clustering. 
Therefore, the source-guided target alignment loss is given by:

\begin{equation}
    \mathcal{L}_{\textrm{TA}} = \sum_{i=1}^{3N_k} \frac{-1}{|S_{-i}|} \sum_{j\in S_{-i}} \log \left(\frac{\exp(z_i . z_j/\tau)}{\sum_{k\neq i}\exp(z_i .z_k/\tau)}\right)
\end{equation}


\begin{table*}[!t]
\begin{adjustbox}{max width = \linewidth}
\begin{tabular}{c|l|ccccccccccccccc|c}
\hline
\multicolumn{1}{l|}{Dataset} &
  Method &
  \multicolumn{1}{l}{\rotatebox{70}{gaussian}} &
  \multicolumn{1}{l}{\rotatebox{70}{shot}} &
  \multicolumn{1}{l}{\rotatebox{70}{impulse}} &
  \multicolumn{1}{l}{\rotatebox{70}{defocus}} &
  \multicolumn{1}{l}{\rotatebox{70}{glass}} &
  \multicolumn{1}{l}{\rotatebox{70}{motion}} &
  \multicolumn{1}{l}{\rotatebox{70}{zoom}} &
  \multicolumn{1}{l}{\rotatebox{70}{snow}} &
  \multicolumn{1}{l}{\rotatebox{70}{frost}} &
  \multicolumn{1}{l}{\rotatebox{70}{fog}} &
  \multicolumn{1}{l}{\rotatebox{70}{brightness}} &
  \multicolumn{1}{l}{\rotatebox{70}{contrast}} &
  \multicolumn{1}{l}{\rotatebox{70}{elastic}} &
  \multicolumn{1}{l}{\rotatebox{70}{pixelate}} &
  \multicolumn{1}{l|}{\rotatebox{70}{jpeg}} &
  \multicolumn{1}{l}{Mean} \\ \hline
\multirow{6}{*}{\rotatebox{90}{\;\;\small CIFAR-10C}}  & Source & 72.3 & 65.7 & 72.9 & 46.9 & 54.3 & 34.8 & 42   & 25.1 & 41.3 & 26   & 9.3  & 46.7 & 26.6 & 58.5 & 30.3 & 43.5 \\
                            & BN Stats Adapt \cite{bn_neurips} & 28.1 & 26.1 & 36.3 & 12.8 & 35.3 & 14.2 & 12.1 & 17.3 & 17.4 & 15.3 & 8.4  & 12.6 & 23.8 & 19.7 & 27.3 & 20.4 \\
                            & TENT-continual \cite{tent} & 24.8 & 20.6 & 28.6 & 14.4 & 31.1 & 16.5 & 14.1 & 19.1 & 18.6 & 18.6 & 12.2 & 20.3 & 25.7 & 20.8 & 24.9 & 20.7 \\
                            & CoTTA \cite{cotta}        & 24.3 & 21.3 & 26.6 & 11.6 & 27.6 & 12.2 & 10.3 & 14.8 & 14.1 & 12.4 & 7.5  & 10.6 & 18.3 & 13.4 & 17.3 & 16.2 \\ 
                            & \textbf{SATA}        & 23.9 & 20.1 & 28.0 & 11.6 & 27.4 & 12.6 & 10.2 & 14.1 & 13.2 & 12.2 & 7.4 & 10.3 & 19.1 & 13.3 & 18.5 & {\bf 16.1} {\footnotesize $\pm$ 0.06}\\ \hline
\multirow{6}{*}{\rotatebox{90}{\;\;\small CIFAR-100C}} & Source & 73   & 68   & 39.4 & 29.3 & 54.1 & 30.8 & 28.8 & 39.5 & 45.8 & 50.3 & 29.5 & 55.1 & 37.2 & 74.7 & 41.2 & 46.4 \\
                            & BN Stats Adapt & 42.1 & 40.7 & 42.7 & 27.6 & 41.9 & 29.7 & 27.9 & 34.9 & 35   & 41.5 & 26.5 & 30.3 & 35.7 & 32.9 & 41.2 & 35.4 \\
                            & TENT-continual & 37.2 & 35.8 & 41.7 & 37.9 & 51.2 & 48.3 & 48.5 & 58.4 & 63.7 & 71.1 & 70.4 & 82.3 & 88   & 88.5 & 90.4 & 60.9 \\
                            & CoTTA          & 40.1 & 37.7 & 39.7 & 26.9 & 38   & 27.9 & 26.4 & 32.8 & 31.8 & 40.3 & 24.7 & 26.9 & 32.5 & 28.3 & 33.5 & 32.5 \\ 
                            & \textbf{SATA}         & 36.5 & 33.1 & 35.1 & 25.9 & 34.9 & 27.7 & 25.4 & 29.5 & 29.9 & 33.1 & 23.6 & 26.7 & 31.9 & 27.5 & 35.2 & {\bf 30.3} {\footnotesize $\pm$ 0.05} \\ \hline
\multirow{6}{*}{\rotatebox{90}{\;\;\small ImageNet-C}} & Source & 97.8 & 97.1 & 98.2 & 81.7 & 89.8 & 85.2 & 78   & 83.5 & 77.1 & 75.9 & 41.3 & 94.5 & 82.5 & 79.3 & 68.6 & 82   \\
                            & BN Stats Adapt & 85   & 83.7 & 85   & 84.7 & 84.3 & 73.7 & 61.2 & 66   & 68.2 & 52.1 & 34.9 & 82.7 & 55.9 & 51.3 & 59.8 & 68.6 \\
                            & TENT-continual & 81.6 & 74.6 & 72.7 & 77.6 & 73.8 & 65.5 & 55.3 & 61.6 & 63   & 51.7 & 38.2 & 72.1 & 50.8 & 47.4 & 53.3 & 62.6 \\
                            & CoTTA          & 84.7 & 82.1 & 80.6 & 81.3 & 79   & 68.6 & 57.5 & 60.3 & 60.5 & 48.3 & 36.6 & 66.1 & 47.2 & 41.2 & 46   & 62.7 \\ 
                            & \textbf{SATA}         & 74.1 & 72.9 & 71.6 & 75.7 & 74.1 & 64.2 & 55.5 & 55.6 & 62.9 & 46.6 & 36.1 & 69.9 & 50.6 & 44.3 & 48.5 & {\bf 60.1} {\footnotesize $\pm$ 0.06} \\ \hline
\end{tabular}
\end{adjustbox}
\caption{Error percentages (lower is better) of different algorithms for CIFAR-10C, CIFAR-100C and ImageNet-C for batchsizes of 200, 200 and 64 respectively. For SATA the standard deviation is reported over 5 random seeds.}
\label{table:comparison}
\end{table*}

\subsection{Final loss}
Given any off-the-shelf pre-trained model and the source prototypes, during testing, using the current test batch, we modify the BN parameters such that the following loss is minimized:
\begin{equation}
    \mathcal{L_{\textrm{SATA}}} = \mathcal{L}_{\textrm{SA}} + \mathcal{L}_{\textrm{TA}}
\end{equation}
The adapting model is used for predicting the class of all the test samples. 
It is robust enough to be updated continuously without any restoration back to the source model.
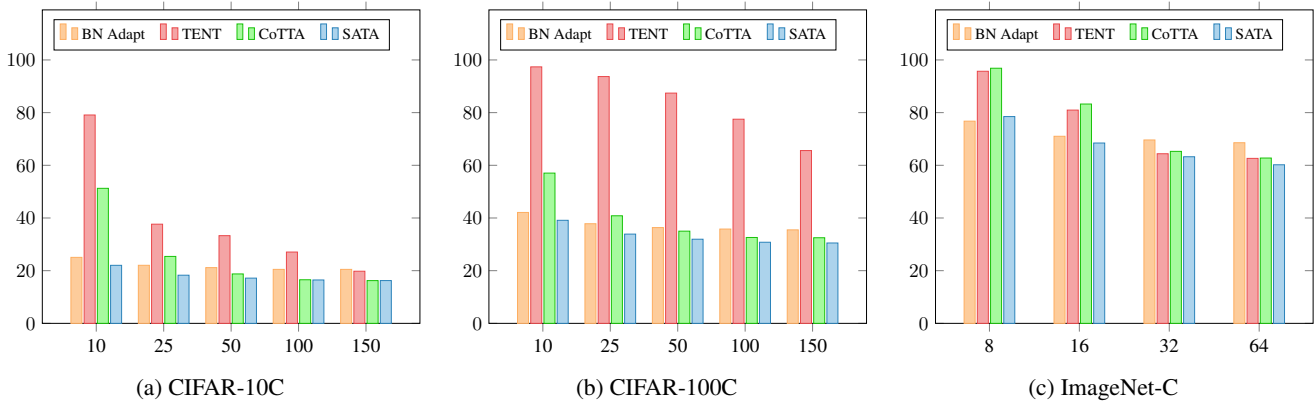
\begin{figure*}[t!]
     \centering
     \begin{subfigure}[b]{0.32\linewidth}
         \centering
                  \begin{adjustbox}{max width=\linewidth}
         \begin{tikzpicture}
            \pgfplotstableread{
                X       a   b   c   d
                10       25.05	79.10	51.26	22.03
                25       22.04	37.65	25.41	18.28
                50       21.16	33.29	18.76	17.17
                100      20.49	27.07	16.55	16.47
                150      20.49	19.79	16.23	16.24
                             }\mydata
            \begin{axis}[
                ybar=0.4mm,     
                bar width=2mm,  
                ymin=0, ymax=119,
            %
                xtick=data,
                xticklabels from table = {\mydata}{X},
                enlarge x limits = 0.2,
            %
                legend style = {legend columns=-1,
                                legend pos=north west,
                                font=\footnotesize,
                                /tikz/every even column/.append style={column sep=2mm},
                                },
                ]
            \addplot[yellow1, fill=yellow-light] table[x expr=\coordindex,y index=1] {\mydata};
            \addplot[red1, fill=red-light] table[x expr=\coordindex,y index=2] {\mydata};
            \addplot[green1, fill=green-light] table[x expr=\coordindex,y index=3] {\mydata};
            \addplot[blue1, fill=blue-light] table[x expr=\coordindex,y index=4] {\mydata};
            
            \legend{BN Adapt, TENT, CoTTA, SATA}
            
            \end{axis}
            \end{tikzpicture}
         \end{adjustbox}
         \caption{CIFAR-10C}
     \end{subfigure}
     \hfill
     \begin{subfigure}[b]{0.32\linewidth}
         \centering
         \begin{adjustbox}{max width=\linewidth}
         \begin{tikzpicture}
            \pgfplotstableread{
                X       a   b   c   d
                10       42.10	97.37	57.04	39.12
                25       37.82	93.69	40.84	33.88
                50       36.36	87.41	35.00	31.94
                100      35.80	77.50	32.60	30.79
                150      35.49	65.60	32.50	30.51
                             }\mydata
            \begin{axis}[
                ybar=0.4mm,     
                bar width=2mm,  
                ymin=0, ymax=119,
            %
                xtick=data,
                xticklabels from table = {\mydata}{X},
                enlarge x limits = 0.2,
            %
                legend style = {legend columns=-1,
                                legend pos=north west,
                                font=\footnotesize,
                                /tikz/every even column/.append style={column sep=2mm},
                                },
                ]
            \addplot[yellow1, fill=yellow-light] table[x expr=\coordindex,y index=1] {\mydata};
            \addplot[red1, fill=red-light] table[x expr=\coordindex,y index=2] {\mydata};
            \addplot[green1, fill=green-light] table[x expr=\coordindex,y index=3] {\mydata};
            \addplot[blue1, fill=blue-light] table[x expr=\coordindex,y index=4] {\mydata};
            
            \legend{BN Adapt, TENT, CoTTA, SATA}
            
            \end{axis}
            \end{tikzpicture}
         \end{adjustbox}
         \caption{CIFAR-100C}
     \end{subfigure}
     \hfill
     \begin{subfigure}[b]{0.32\linewidth}
         \centering
         \begin{adjustbox}{max width=\linewidth}
         \begin{tikzpicture}
            \pgfplotstableread{
                X       a   b   c   d
                8       76.76	95.7	96.85	78.5
                16       71.02	80.96	83.26	68.46
                32       69.62	64.39	65.29	63.25
                64      68.57	62.63	62.76	60.2
                             }\mydata
            \begin{axis}[
                ybar=0.4mm,     
                bar width=2mm,  
                ymin=0, ymax=119,
            %
                xtick=data,
                xticklabels from table = {\mydata}{X},
                enlarge x limits = 0.2,
            %
                legend style = {legend columns=-1,
                                legend pos=north west,
                                font=\footnotesize,
                                /tikz/every even column/.append style={column sep=2mm},
                                },
                ]
            \addplot[yellow1, fill=yellow-light] table[x expr=\coordindex,y index=1] {\mydata};
            \addplot[red1, fill=red-light] table[x expr=\coordindex,y index=2] {\mydata};
            \addplot[green1, fill=green-light] table[x expr=\coordindex,y index=3] {\mydata};
            \addplot[blue1, fill=blue-light] table[x expr=\coordindex,y index=4] {\mydata};
            
            \legend{BN Adapt, TENT, CoTTA, SATA}
            
            \end{axis}
            \end{tikzpicture}
         \end{adjustbox}
         \caption{ImageNet-C}
     \end{subfigure}
    \caption{These figures compare the robustness of the different approaches for different batch sizes. The x-axis is the batch size and y-axis is the error percentage (lower is better). We observe that the proposed SATA is robust across batch sizes.}
    \label{fig:low-batch}
\end{figure*}

\section{Experimental Evaluation} 
Here we describe the extensive experiments performed to evaluate the effectiveness of the proposed framework. \\ 
\textbf{Dataset Details}: Here, we evaluate the proposed framework extensively on multiple benchmark datasets, namely, CIFAR-10C, CIFAR-100C and ImageNet-C~\cite{imagenetc}. These datasets have 10, 100 and 1000 classes respectively.
All the datasets contain $15$ diverse forms of corruption (noise, blur, weather, and digital) with five levels of severity, applied to the test set of all the three datasets.
For all the experiments, unless mentioned otherwise, the test sequence consists of all $15$ corruptions at the highest level of severity~\cite{cotta}.
The goal is to adapt an off-the-shelf source model to this dynamically changing environment efficiently during test time. \\ \\
\textbf{Research Questions:} The research questions that we want to answer using the experiments are the following: \\
i) How does the proposed framework perform on these  datasets using the standard experimental protocol (higher batch size) as used in~\cite{cotta}? \\
ii) How does the model perform with lower batch sizes, which are more realistic in an online setting as in~\cite{tent}? \\
iii) Are we able to learn a generalized model which also retains its good performance on the data from source distribution? \\
iv) Are both the proposed modules important? \\
iv) Practical considerations - How does the model fare in terms of storage cost, inference time, number of hyperparameters to be tuned across datasets? \\ \\
\textbf{Implementation Details}: 
For CIFAR-10C, we use a pre-trained WideResNet-28 \cite{wideresnet} model from the RobustBench benchmark \cite{robustbench} as in~\cite{cotta}. 
The model is updated with one gradient step per iteration, and the Adam optimizer with a learning rate of 1e-3 is used. The temperature is set to the default value of 0.1.
The CIFAR-100C experiment uses a pre-trained ResNeXt-29 \cite{resnext} model, which is one of the default models for CIFAR-100 in the RobustBench benchmark \cite{robustbench}. The same hyperparameters as the CIFAR-10 experiment are used.
For the ImageNet-C experiment, the standard pre-trained Resnet50 \cite{resnet} model from RobustBench \cite{robustbench} is used. Here, SGD is used as the optimiser with a learning rate of 1e-2 as in \cite{cotta}. We conduct all the experiments on an NVIDIA GeForce RTX 3090.
\subsection{Evaluation on Standard Benchmarks}
Table \ref{table:comparison} reports the results on the three  benckmark datasets.
The batch sizes used for these experiments are $200$, $200$ and $64$ for CIFAR-10C, CIFAR-100C and ImageNet-C respectively as in~\cite{cotta}.
All the results for the other approaches are directly taken from \cite{cotta}.
In the TENT-continual setup, the model continuously adapts and is not reset to the source model after each corruption.
We observe that for all the datasets, the proposed SATA outperforms all the other existing approaches.
Specifically, for the challenging ImageNet-C, we obtain an error of $60.1\%$, which is $2.6\%$ better than the previous state-of-the-art CoTTA.
In addition to the gain in performance, SATA has other advantages, as elaborated on later.

\begin{table*}
\begin{adjustbox}{max width=\linewidth}
\begin{tabular}{l|cccccc|cccccc|cccc}
\hline
\multicolumn{1}{c|}{\multirow{2}{*}{Methods}} & \multicolumn{6}{c|}{CIFAR-10C} & \multicolumn{6}{c|}{CIFAR-100C} & \multicolumn{4}{c}{IMAGENET-C} \\ \cline{2-17} 
\multicolumn{1}{c|}{} & 200 & 150 & 100 & 50 & 25 & 10 & 200 & 150 & 100 & 50 & 25 & 10 & 64 & 32 & 16  & 8 \\ 
\hline
SALoss (original image) \color{black} & 20.1 & 20.1 & 20.4 & 20.7 & 21.7 & 24.8 & 32.6 & 32.8 & 33.2 & 34.0 & 35.7 & 40.7 & 62.7 & 64.9 & 69.4 & 77.8 \\
SALoss (original + augmented image) & 17.1 & 17.2 & 17.4 & 18.0 & 19.0 & 22.5 & 31.4 & 31.6 & 32.0 & 32.8 & 34.7 & 39.9 & 61.3 & 63.8 & 68.5 & {\bf 76.8} \\
{\bf SATA} (SALoss + TALoss) & {\bf 16.1} & {\bf 16.2} & {\bf 16.5} & {\bf 17.2} & {\bf 18.3} & {\bf 22.0} & {\bf 30.4} & {\bf 30.5} & {\bf 30.8} & {\bf 31.9} & {\bf 33.9} & {\bf 39.1} & {\bf 60.2} & {\bf 63.2} & {\bf 68.5} & 78.5 \\ 
\hline
\end{tabular}
\end{adjustbox}
\caption{Mean error percentage (lower is better) demonstrating the importance of the two proposed components. 
Ablation is done on all the four datasets for multiple batch sizes. 
}
\label{table:ablation}
\end{table*}

\subsection{Evaluation with Lower Batch Sizes}
A practical test-time adaptation algorithm should work satisfactorily for lower batch sizes, which will decrease the average time for inference of a sample and thus lower the latency of the framework. 
Recently, researchers have started to address the issue of robustness across batch sizes~\cite{ttn}, but many of these methods are not fully test time adaptation (FTTA) and requires source training for initialisation.

Here, we evaluate the robustness of the proposed SATA framework for lower batch sizes and compare the results with the state-of-the-art. 
Figure~\ref{fig:low-batch} shows the results for the three datasets with decreasing batch sizes.
Since the results of the other approaches were not reported for other batch sizes, we ran the official codes and obtained the results reported in the table. To ensure the best results for the other methods, we tuned the appropriate hyperparameters. 
Specifically, for TENT, we varied the learning rate. 
For CoTTA, the restoration probability and model EMA factor was reduced in proportion to the decrease in batch size. 
Changing any other parameters did not substantially change the performance of the model.
Note that for the proposed SATA, {\em no parameters were changed across the different datasets as well as batch sizes}. 
We observe that the improvement provided by our approach becomes clearer as the batch size decreases.  For example, for a batch size of 10, we obtain 22.0\% for CIFAR-10C dataset, which is significantly better compared to the next best obtained by BNStats.
Though there is still a lot of room for improvement for all the frameworks, this experiment justifies the effectiveness of the proposed SATA for online setting. 
\subsection{Ablation Study}  Here, we analyze the importance of the two losses in the proposed SATA framework.
We observe from Table~\ref{table:ablation} that most of the performance improvement of SATA can be attributed to the source-anchoring of the test samples and its augmented version.
The clustering and alignment terms further help to improve the performance, thereby achieving state-of-the-art performance for continual test-time domain adaptation under different challenging settings.





\section{Further Analysis}
Here, we perform further analysis to evaluate the usefulness of the proposed framework and its different components.
All these analysis are done on the CIFAR-100C datasets, unless stated otherwise. \\ \\
\textbf{Performance on gradually changing data:} In the standard setup, the corruption types change abruptly with maximum severity levels.
A more realistic approach will be to evaluate the performance of the approaches when the severity levels change gradually over a sequence of 15 corruption types. Thus we experiment with the gradual setup as also done in~\cite{cotta}. The representation below shows the order in which severity is faced for every corruption. \\
\resizebox{\linewidth}{!}{
{\scriptsize 
$\underbrace{\hdots\rightarrow2\rightarrow1}_\text{\scriptsize {t-1 and before}} \rightarrow \underbrace{1\rightarrow2\rightarrow3\rightarrow4\rightarrow5\rightarrow4\rightarrow3\rightarrow2\rightarrow1}_{\text{\scriptsize t corruption type, with changing severity}} \rightarrow \underbrace{1\rightarrow2\rightarrow\hdots}_\text{\scriptsize {t+1 and after}} 
$}}
\\
Table \ref{table:gradual} reports the results of the proposed framework and comparisons with the existing approaches for this setup. The results suggest that BN Adapt, CoTTA, and SATA are more effective than the source and TENT (cont) approaches in dealing with corruption types that change gradually over time. In particular, our approach achieves the lowest average error rate of 25.6\%, followed by the CoTTA approach with an average error rate of 26.3\%. The BN Adapt approach also performs well with an average error rate of 29.9\%.

\begin{table}[]
\begin{adjustbox}{max width=\linewidth}
\begin{tabular}{@{}llllll@{}}
\toprule
Avg. Error (\%) & Source & BN Adapt & TENT (cont) & CoTTA & SATA \\ \midrule
CIFAR-100C      & 33.6   & 29.9     & 74.8        & 26.3  & {\bf 25.6}  \\
\bottomrule
\end{tabular}
\end{adjustbox}
\caption{Average error over all corruptions with severity presented in the gradual test time adaptation manner. The order of corruption used here is the same as in Table~\ref{table:comparison}. }
\label{table:gradual}
\end{table}
\begin{table}[t!]
\begin{adjustbox}{max width = \linewidth}
\begin{tabular}{@{}lllllll@{}}
\toprule
batchsize $\rightarrow$  & 200  & 150  & 100  & 50   & 25   & 10   \\ 
\midrule
TENT  & 92.0 (70.9) & 96.2 (75.1) & 97.8 (76.7) & 98.2 (77.1) & 98.5 (77.4)  & 99.0 (77.9) \\
CoTTA & 22.8 (1.7) & 23.0 (1.9)  & 23.9 (2.8) & 27.2 (6.1) & 35.5 (14.4) & 58.5 (37.4) \\ 
SATA  & \textbf{22.4 (1.3)} & {\bf 22.7 (1.6)} & {\bf 22.9 (1.8)} & {\bf 23.6 (2.5)}  & {\bf 25.8 (4.7)} & \textbf{29.9 (8.8)} \\ 
\bottomrule
\end{tabular}
\end{adjustbox}
\caption{Error on CIFAR-100 test set after the model has been adapted to all the corruptions as in Table~\ref{table:comparison}. $(.)$ is the degradation in performance on source data compared to the source model, whose error is 21.1\%. 
This deviation can be thought of as a proxy for catastrophic forgetting. 
}
\label{tab:adapt-scr}
\end{table}

%
\noindent{\bf Performance on source data: }
As the trained model gradually adapts to the changing testing conditions, we want it to be able to perform well on the original source distribution, which requires that the model has not catastrophically forgotten the original training information. 
Eg, the original model trained on clear weather conditions should continue to do well for clear weather images, even though it has adapted to other conditions like rainy, foggy, etc. 

To achieve these contrasting goals, it is important that the model has the right balance of stability-plasticity. The stability-plasticity trade-off refers to the balance between preserving learned knowledge and adapting to new information. 
Table \ref{tab:adapt-scr} reports the results of using the adapted model on a held-out testing set from the source distribution on CIFAR-100C dataset.
The performance of the original model trained on source gives an error of $21.1\%$ on its test set. Thus the difference gives an estimate of the catastrophic forgetting (Table ~\ref{tab:adapt-scr}).
We observe that even after adaptation, SATA is able to maintain its performance on the source distribution very well as compared to CoTTA and TENT for all batch sizes. \\ \\
\textbf{Generalizability of the learnt model:} When the source model encounters data from different domains, we want it to gradually become more generalized, such that the features become domain invariant.
To evaluate whether this happens in practice, we perform an experiment, in which the model is first adapted on the first 7 corruptions (gaussian to zoom).
We then freeze the weights and evaluate its performance on the last 8 corruptions (snow to jpeg). 
We compare the performance of this adapted model (adapted using CoTTA and SATA) to the performance using the source model, whose BN statistics are adapted to the corresponding batch statistics.  
We see from the results in Table~\ref{table:generalization} that both CoTTA and the proposed framework have indeed learnt more generalized features and is therefore performing better than the source model (BN Adapt).
To understand the generalization capability provided by our individual losses, we also report the results using only the source-anchoring loss ($\mathcal{L}_{\textrm{SA}}$).
We observe that this single loss term gives comparable performance as CoTTA, even without the target alignment loss. \\ 

\begin{table}[]
\begin{adjustbox}{max width = \linewidth}
\begin{tabular}{@{}lccccccccl@{}}
\toprule
  \multicolumn{1}{l}{Method} &
  \multicolumn{1}{l}{\rotatebox{70}{snow}} &
  \multicolumn{1}{l}{\rotatebox{70}{frost}} &
  \multicolumn{1}{l}{\rotatebox{70}{fog}} &
  \multicolumn{1}{l}{\rotatebox{70}{brightness}} &
  \multicolumn{1}{l}{\rotatebox{70}{contrast}} &
  \multicolumn{1}{l}{\rotatebox{70}{elastic}} &
  \multicolumn{1}{l}{\rotatebox{70}{pixelate}} &
  \multicolumn{1}{l}{\rotatebox{70}{jpeg}} &
  \multicolumn{1}{l}{Mean} \\ \midrule
BN Adapt & 35.6 & 34.9 & 42.1 & 26.9 & 31.0 & 36.0 & 33.5 & 41.7 & 35.2 \\
CoTTA & 31.3 &	30.5&	36.7&	25.2&	27.8&	31.5&	29.1&	36.4&	\textbf{31.1} \\
\midrule
{SALoss} & 31.9&	31.3&	37.2	&25.0	&29.4&	33.0&	29.2	&37.8&	31.9 \\
SATA & 31.2 & 31.2 & 36.3 & 25.0 & 28.7 & 32.2 & 28.0 & 37.2 & 31.2 \\ \bottomrule
\end{tabular}
\end{adjustbox}
\caption{This experiment on CIFAR-100C demonstrates the generalizability of the learnt model. We observe that both CoTTA and SATA yield a more generalized model after adaptation.}
\label{table:generalization}
\end{table}

\noindent\textbf{Effect of source prototypes:} In some cases, prototypes may not be available. We check the performance of our method in such cases as demonstrated in Table \ref{tab:con-ab}. We observe that without the prototypes, the performance drops slightly compared to our complete loss. This drop is small because the other loss component ($\mathcal{L}_\text{SA}$) will increase if the features of the adapting model are not aligned with the features of the source. Thus, even if only the pre-trained model is available, our method can be used for test time adaptation.

\begin{table}[h]
\begin{adjustbox}{max width = \linewidth}
\begin{tabular}{@{}llll@{}}
\toprule
                    & CIFAR-10C   & CIFAR-100C  & ImageNet-C  \\ 
\midrule
SATA w/ prototypes  & 16.1        & 30.0        & 60.2          \\ 
SATA w/o prototypes & 16.2 (-0.1) & 31.4 (-1.4) & 61.1 (-0.9)   \\
\bottomrule
\end{tabular}
\end{adjustbox}
\caption{Here, we see the performance comparison (error \% - lower is better) between using prototypes as a view v/s not using them in the target alignment loss. 
(.) is the drop in performance compared to the complete loss.}
\label{tab:con-ab}
\end{table}

\noindent \textbf{Computational Advantages}: 
The proposed SATA framework has additional advantages over the existing approaches as follows:\\
1) Less memory requirements due to less number of trainable parameters;\\
2) Faster inference time due to lesser number of forward passes. \\
{\em \textbf{Number of parameters to be stored}}:
Table \ref{table:num_params} reports the total number of parameters and trainable parameters for TENT, CoTTA and SATA. 
We observe that CoTTA has 33\% more parameters than SATA and the trainable parameters in SATA is only 2.1\% of those in CoTTA.
Thus the proposed framework is much simpler and computationally efficient, which can be especially important in real-time applications where time and computational resources are limited. 
\begin{table}[h]
\centering
\begin{adjustbox}{max width=0.7\linewidth}
\begin{tabular}{@{}llllll@{}}
\toprule
Method     & \# Parameters & \# Trainable & \% Trainable \\ 
\midrule
BN-Stats & 6,900,132 & 0    & 0     \\ 
TENT     & 6,900,132 & 128  & 0.002 \\ 
CoTTA & 20,700,396   & 6,900,132   & 33.333 \\ 
\midrule
SATA    & 13,947,976 & 147,840     & 1.060  \\ 
\bottomrule
\end{tabular}
\end{adjustbox}
\caption{Number of (trainable) parameters as a proxy for the storage requirement of the respective algorithms. 
This table is for CIFAR-100C with ResNeXt-29 as the backbone.}
\label{table:num_params}
\end{table}

\begin{figure}
\centering
\begin{adjustbox}{max width=\linewidth}
\begin{tikzpicture}
\begin{axis}[
    xbar stacked,
    legend style={
        legend columns=4,
        at={(xticklabel cs:0.5)},
        anchor=north,
        draw=none
    },
    ytick=data,
    axis y line*=none,
    axis x line*=bottom,
    tick label style={font=\footnotesize},
    legend style={font=\footnotesize},
    label style={font=\footnotesize},
    xtick={0,0.5,1.0,1.5,2.0,2.5},
    width=.8\textwidth,
    bar width=6mm,
    xlabel=Time,
    yticklabels={CoTTA, SATA, SATA},
    xmin=0,
    xmax=2.5,
    area legend,
    y=10mm,
    enlarge y limits={abs=0.625},
]
\addplot[blue1,fill=blue1] coordinates
{ (1.775,1) (0.2,2)}; 
\addplot[red1,fill=red1] coordinates
{ (0.0539,1)(0.021,2)}; 
\addplot[yellow1,fill=yellow1] coordinates
{ (0.385,1)(0,2)}; 

\legend{Forward Pass, Backward optimisation, Post Backward}
\end{axis}
\end{tikzpicture}
\end{adjustbox}
\caption{Comparison of inference time of the proposed SATA framework with the state-of-the-art CoTTA. 
The x-axis is time of inference per batch (sec/batch). These experiments were done on CIFAR-100C using NVIDIA GeForce RTX 3090.}
\label{fig:inf-time}
\end{figure}
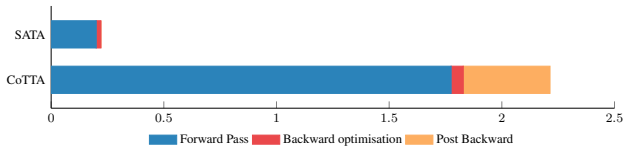

\noindent {\em \textbf{Inference time for a fixed batch size}}:
The inference time of a model is an important factor in test-time adaptation because it determines how quickly the model can be applied to new data. In Figure \ref{fig:inf-time} we have compared the inference time (in sec) per batch (of 200 samples) for CoTTA and SATA. 
We see SATA is around 10 times faster than CoTTA during inference. 
This can be attributed to the fact that CoTTA uses 32 forward passes in the worst case for its prediction and also updates the teacher model after every step. 

\section{Conclusion}

In this paper, we proposed a novel SATA framework for the challenging task of continual test-time domain adaptation.
The proposed approach modifies the batch-norm affine parameters using source anchoring-based self-distillation to ensure the model incorporates knowledge of newly encountered domains while avoiding catastrophic forgetting. 
Additionally, source prototype guided target alignment is proposed to maintain the already learned semantic information while grouping target samples naturally. 
The approach is quite robust to decreasing batch sizes, justifying its effectiveness for online application. 
But we observe that for very small batch sizes (eg 8 in ImageNet), its performance drops slightly below BN Adapt, though even for this case, it is significantly better than TENT and CoTTA. 
The SATA framework offers additional advantages like retaining performance on the source domain, and having minimal tunable hyper-parameters and storage requirements, in addition to achieving state-of-the-art results on all the benchmark datasets.

{\small
\bibliographystyle{ieee_fullname}
\bibliography{bib}
}

\end{document}